\documentclass[lettersize,journal]{IEEEtran}
\usepackage{amsmath,amsfonts}
\usepackage{algorithmic}
\usepackage{algorithm}
\usepackage{array}
\usepackage[caption=false,font=normalsize,labelfont=sf,textfont=sf]{subfig}
\usepackage{textcomp}
\usepackage{stfloats}
\usepackage{url}
\usepackage{verbatim}
\usepackage{graphicx}
\usepackage{cite}
\usepackage{booktabs}
\usepackage{algorithm}
\usepackage{amsmath}
\usepackage{amssymb}
\usepackage{mathtools}
\usepackage{amsthm}
\usepackage{multirow}

\begin{document}
	
	\newtheorem{define}{Lemma}
	
	
	\title{Uncertainty-aware Long-tailed Weights Model the Utility of Pseudo-labels for Semi-supervised Learning}
	
	\author{Jiaqi Wu, Junbiao Pang, Qingming Huang,~\IEEEmembership{Fellow,~IEEE}}
	
	\maketitle
	
	\begin{abstract}
		Current Semi-supervised Learning (SSL) adopts the pseudo-labeling strategy and further filters pseudo-labels based on confidence thresholds. However, this mechanism has notable drawbacks: 1) setting the reasonable threshold is an open problem which significantly influences the selection of the high-quality pseudo-labels; and 2) deep models often exhibit the over-confidence phenomenon which makes the confidence value an unreliable indicator for assessing the quality of pseudo-labels due to the scarcity of labeled data. In this paper, we propose an Uncertainty-aware Ensemble Structure (UES) to assess the utility of pseudo-labels for unlabeled samples. We further model the utility of pseudo-labels as long-tailed weights to avoid the open problem of setting the threshold. Concretely, the advantage of the long-tailed weights ensures that even unreliable pseudo-labels still contribute to enhancing the model's robustness. Besides, UES is lightweight and architecture-agnostic, easily extending to various computer vision tasks, including classification and regression. Experimental results demonstrate that combining the proposed method with DualPose leads to a 3.47\% improvement in Percentage of Correct Keypoints (PCK) on the Sniffing dataset with 100 data points (30 labeled), a 7.29\% improvement in PCK on the FLIC dataset with 100 data points (50 labeled), and a 3.91\% improvement in PCK on the LSP dataset with 200 data points (100 labeled). Furthermore, when combined with FixMatch, the proposed method achieves a 0.2\% accuracy improvement on the CIFAR-10 dataset with 40 labeled data points and a 0.26\% accuracy improvement on the CIFAR-100 dataset with 400 labeled data points. Code is publicly available at: \url{https://github.com/Qi2019KB/UES}
	\end{abstract}
	
	
	\section{Introduction}
	\label{sec:Introduction}
	
	The state-of-the-art (SOTA) semi-supervised learning (SSL) methods, such as FixMatch~\cite{sohn2020fixmatch} and FreeMatch~\cite{wang2022freematch}, adopt the pseudo-labeling strategy~\cite{zhang2021flexmatch, huang2023flatmatch, jiang2022maxmatch, xie2021empirical}. These methods center on a strong and weak data augmentation framework. For unlabeled data, the model generates predictions through weak augmentation. If the prediction confidence exceeds a preset threshold, the pseudo-label is used for training, also known as the Self-Training Strategy (STS). STS assumes that high-confidence pseudo-labels typically indicate high quality, reducing noise interference during training.
	
	However, relying on confidence thresholds to filter pseudo-labels presents three main issues. Firstly, setting a reasonable threshold is challenging, as it must effectively select high-quality pseudo-labels. A low threshold introduces noise, while a high threshold limits the number of pseudo-labels. For instance, FixMatch trains with labeled data until confidences exceed the threshold, risking overfitting and exacerbating the Matthew Effect, where accurate classes improve further, while inaccurate ones degrade.
	
	Secondly, the confidence threshold approach may discard valuable supervisory information. As noted in~\cite{grandvalet2004semi}, unlabeled examples are most beneficial when classes have small overlap. CBE~\cite{wu2024channel} adds that, in early SSL stages, even incorrect pseudo-labels may contain useful information. However, the confidence threshold inherently disregards such potentially valuable information.
	
	Furthermore, due to limited labeled data, models can be overconfident~\cite{yao2023pseudo, ling2022semi, chen2024self}, making confidence an unreliable indicator of pseudo-label accuracy. Confidence reflects model certainty, not prediction accuracy. Thus, the confidence threshold may introduce noisy labels, affecting training effectiveness. This is particularly evident in tasks like pose estimation, where confidence only indicates certainty about keypoint locations on heatmaps.
	
	This paper proposes a lightweight, architecture-agnostic method, Uncertainty-aware Ensemble Structure (UES), to assess pseudo-label utility for unlabeled samples. UES models pseudo-label utility as long-tailed weights, avoiding the threshold setting problem. The advantage of long-tailed weights ensures that even unreliable pseudo-labels contribute to enhancing model robustness. UES is based on two key insights:
	
	\begin{itemize}
		\item \textbf{Mean-based samples uncertainty}: Averaging predictions from multiple base models provides a reasonable reference for assessing uncertainty. The advantage of using a mean-based sample uncertainty method is that it can effectively reduce the adverse effects of excessive prediction deviations from individual prediction heads on the quantification of uncertainty through averaging predictions.
		
		\item \textbf{Prediction head uncertainty}: The uncertainty of a prediction head can be evaluated by calculating the average uncertainty of its outputs across a batch of samples. Furthermore, converting the uncertainty of the prediction head into weights and using a weighted average approach to calculate ensemble predictions effectively prevents overfitting to individual samples and produces more accurate ensemble prediction results than simple averaging.
	\end{itemize}
	
	UES is compatible with classification and regression tasks and can be integrated into SSL frameworks like FixMatch and DualPose. Experiments on pose estimation (Sniffing, FLIC, LSP) and classification (CIFAR-10/100) datasets showed significant improvements. Our contributions are as follows:
	
	\begin{itemize}
		\item Proposing UES, which quantifies sample uncertainty and uses it to guide the loss function calculation. By computing prediction head uncertainty and using it in ensemble prediction, UES enhances accuracy. It is lightweight and applicable to various SSL tasks.
		\item The long-tail weights avoid potential issues that may arise from relying solely on confidence thresholds for evaluating and selecting pseudo-labels, such as complex threshold settings, loss of valuable supervisory information, and susceptibility to noisy pseudo-label selection results due to overfitting.
		
		\item Combining SSL and UES effectively enhances model training performance. In SSL regression tasks, UES improved prediction accuracy by 3.47\%, 7.29\%, and 3.91\% on Sniffing, FLIC, and LSP datasets, respectively, compared to CBE~\cite{wu2024channel}. In SSL classification tasks, UES surpassed baseline models, achieving 0.2\% and 0.26\% increases in prediction accuracy on CIFAR-10 and CIFAR-100 datasets compared to CBE, respectively.
	\end{itemize}
	
	\section{Related Work}
	\label{sec:Related_Work}
	
	\subsection{Quality Evaluation of Pseudo-labels}
	\label{ssec:Quality_Evaluation_of_PLs}
	
	\textbf{Confidence threshold methods}~\cite{sohn2020fixmatch, zhang2021flexmatch, wang2022freematch} retained only the pseudo-labels with prediction confidence higher than a preset threshold. The advantage of this method lies in its simplicity and intuitiveness, making it easy to implement. However, it also had drawbacks: First, the selection of the threshold was subjective, and different threshold settings might lead to vastly different sets of pseudo-labels; Second, in cases of uneven data distribution, some valuable information might be omitted due to the rigid threshold criterion. 
	
	\textbf{Entropy minimization methods}~\cite{grandvalet2004semi, berthelot2019mixmatch, ma2022context, wu2021semi, saito2019semi} aimed to encourage the model to produce low-entropy (i.e., high-certainty) predictions for unlabeled samples. In SSL, these methods typically penalized the model for making high-entropy predictions by adding an entropy regularization term to the loss function for unlabeled samples. This approach prompted the model to make more consistent predictions on unlabeled data, thereby improving the quality of pseudo-labels. However, its drawbacks were that it required additional computational resources to optimize the entropy regularization term, and in some specific scenarios, overly emphasizing low-entropy predictions might lead to model overfitting. 
	
	\textbf{Uncertainty-based methods}~\cite{gal2016dropout, kendall2017uncertainties, wang2021combating, zhao2020uncertainty, rizve2021defense} determined whether to use prediction results as pseudo-labels and further ascertained the weights of pseudo-labels by evaluating the uncertainty of the model's predictions on unlabeled samples. There were various means of quantifying uncertainty. One could precisely calculate prediction uncertainty using techniques such as Bayesian methods or Monte Carlo dropout. Alternatively, one could indirectly obtain it by assessing the consistency of the model's predictions across multiple data augmentation samples. However, it was noteworthy that, under this uncertainty-based evaluation system, low prediction uncertainty did not always equate to high prediction accuracy.
	
	In our UES, we combined uncertainty with confidence to evaluate the pseudo-label weights of unlabeled samples. This approach had the advantage of further eliminating high-confidence erroneous predictions that might arise from model overfitting, by using uncertainty as an additional filter criterion.
	
	\subsection{Quality Evaluation of Base Models in Ensemble Learning}
	\label{ssec:Quality_Evaluation_of_Base_Models}
	
	\textbf{Confidence-based method.~\cite{caruana2004ensemble}} In the realm of deep learning, the confidence scores of individual models for their prediction outcomes served as the basis for weight assignment, where higher confidence translated to heavier weight allocation. This approach offered simplicity and convenience, obviating the need for additional evaluation procedures. Nevertheless, it was crucial to acknowledge that high confidence did not necessarily guarantee prediction accuracy, particularly in scenarios where models were prone to overconfidence, thereby exacerbating the issue.
	
	\textbf{Consistency checking method.~\cite{liang2023label, lu2020mitigating, yang2024segment}} This method indirectly assessed the reliability of different base models by measuring the consistency of their prediction results on the same batch of data. However, it had the following drawbacks: Firstly, it could not directly reflect the accuracy of predictions, as high consistency did not necessarily imply accurate prediction results; all models might have erroneously predicted the same outcome. Secondly, the method was overly sensitive to the diversity of base models; if there were significant differences among the base models, the reliability assessment based on consistency would no longer be accurate.
	
	\textbf{Mean-based uncertainty method.} We propose an uncertainty estimation method based on a simple mean ensemble strategy, inspired by a common observation in the field of ensemble learning: ensemble predictions often outperform individual base models in accuracy. Specifically, this method first calculates the simple average of the prediction results from each base model, using it as the output of the ensemble prediction. Subsequently, it quantifies the uncertainty of the predictions by evaluating the consistency between the predictions of each base model and this ensemble prediction.
	
	Compared to predictions from a single base model, the average of these predictions typically exhibit higher accuracy and stability. Therefore, when using the simple average of predictions of heads as a reference standard, the quantification process of uncertainty is still reliable and robust. Furthermore, this method is applicable to both classification and regression tasks.
	
	\section{Method}
	\label{sec:Method}
	
	\subsection{Overview}
	
	\begin{figure}[h!]
		\centering
		\includegraphics[width=3.5in]{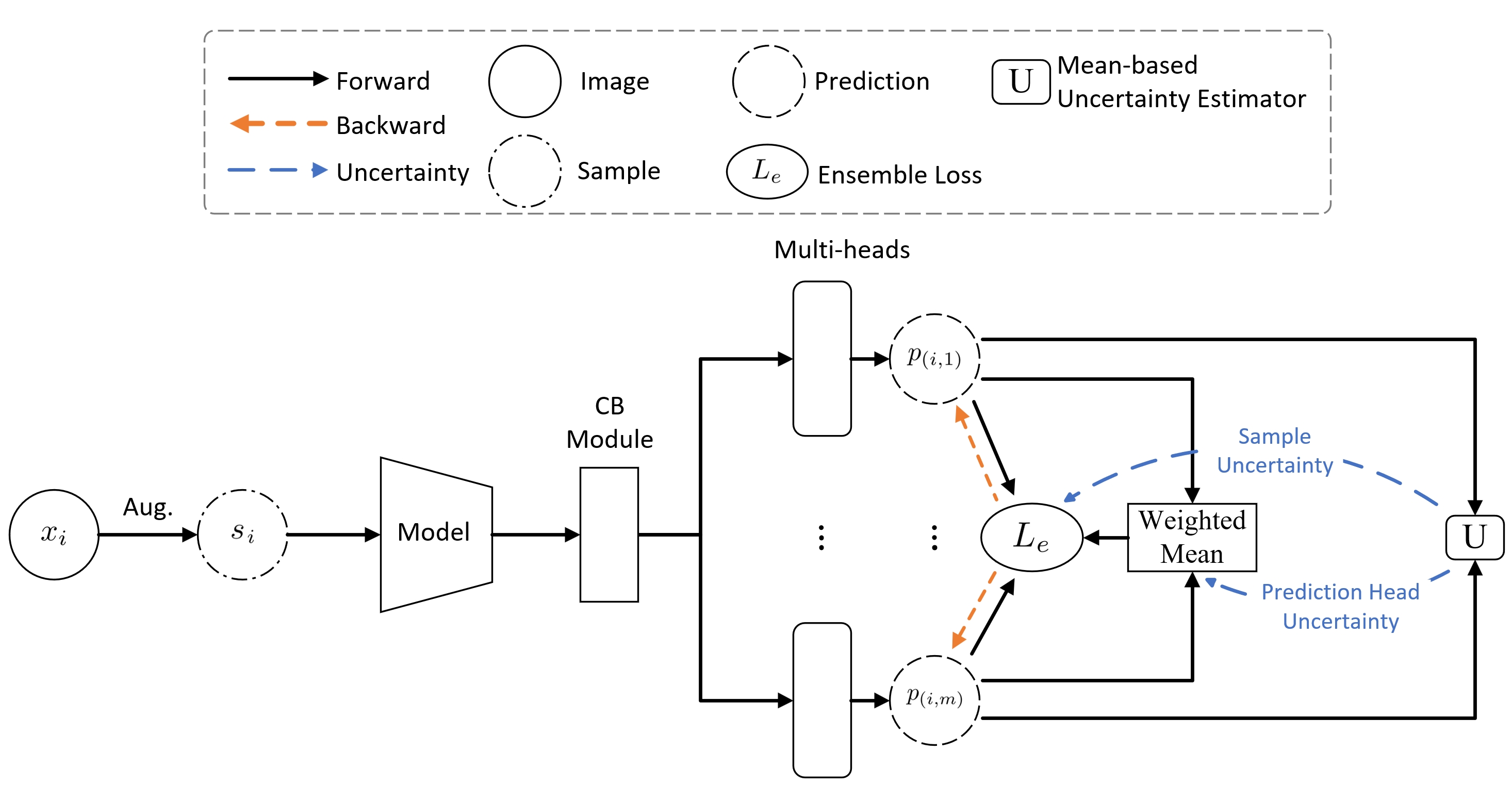}
		\caption{The neural network structure of UES is based on the Channel-based Ensemble (CBE) method~\cite{wu2024decomposed} and a mean-based uncertainty estimator. \label{fig:Network_structure}}
	\end{figure}
	
	The UES network is shown in Fig.~\ref{fig:Network_structure}. UES constructs a lightweight ensemble framework that quantifies sample uncertainty by evaluating the prediction consistency of multiple heads. This uncertainty estimation guides the selection of pseudo-labels in SSL. To achieve this, we propose the Mean-based Uncertainty Estimator (MUE) combined with the Channel-based Ensemble (CBE) method. As a lightweight ensemble framework, CBE effectively achieves high-diversity ensemble predictions without significantly increasing the number of parameters or training time.
	
	Given a batch of unlabeled data $\mathcal{D}_{U}=\{x_{i}\}^{\mu N_B}_{i=1}$, where $N_B$ is the batch size of the labeled data, $\mu$ is the ratio of the unlabeled data number to the labeled data number, each data point $x_i$ in $\mathcal{D}_{U}$ undergoes the following process: 
	\begin{itemize}
		\item First, the unlabeled data $x_i$ is sent into the data augmentation function $\omega(\cdot)$ to generate one data-augmented sample $s_{i}$.
		\item Second, the CB Module of CBE generates $M$ low-correlation features and sends them to corresponding prediction heads to produce $M$ predicted probability distributions $\mathcal{P}_{i}=\{p_{(i,m)}\}^M_{m=1}$.
		\item  Finally, all the predicted distributions $\{\mathcal{P}_{i}\}^{\mu N_B}_{i=1}$ of the batch unlabeled data $\mathcal{D}_{U}$ are sent to the MUE to assess the prediction head uncertainty $\mathcal{U}^{H}$ and the sample uncertainty $\mathcal{U}^{S}$.
	\end{itemize}

	We utilize sample uncertainty $\mathcal{U}^{S}$ to assist in evaluating the credibility of pseudo-labels for unlabeled samples. However, in SSL, due to the lack of label supervision, model predictions on unlabeled samples often suffer from low accuracy and stability, prone to excessive variation among prediction heads and the risk of overfitting to individual samples. Therefore, as illustrated in Fig.~\ref{fig:Unc_corrcoef}, quantifying uncertainty based directly on the prediction consistency of each prediction head is inaccurate.
	
	To address this, we use prediction head uncertainty $\mathcal{U}^{H}$. The assessment of this uncertainty serves as the basis for determining the weights of these prediction heads when calculating the weighted ensemble prediction. The advantages of this approach are:
	
	\begin{itemize}
		\item The uncertainty of prediction heads is derived from their predictions on a batch of samples, which helps mitigate overfitting to individual samples.
		
		\item Through weighted ensemble, we assign higher weights to prediction heads closer to the mean, thereby reducing the adverse effects of excessive prediction variation among prediction heads.
	\end{itemize}
	
	\subsection{Sample Uncertainty}
	\label{ssec:Sample_Uncertainty}
	
	In general, ensemble predictions exhibit a significant improvement in accuracy compared to individual base models. Motivated by this observation, for a batch of samples, we calculate the simple average probability distribution $\widetilde{\mathcal{P}}=\{\widetilde{p}_{i}\}^{\mu N_B}_{i=1}$ from multiple predictions generated by $M$ prediction heads, and use $\widetilde{\mathcal{P}}$ as a reference distribution for computing uncertainty. Here, $\widetilde{p}_{i}=\frac{1}{M}\sum_{m=1}^{M}p_{(i,m)}$, where $N_B$ denotes the batch size of the labeled data, and $\mu$ represents the ratio of the unlabeled data number to the labeled data number.
	
	\textbf{In regression} tasks, for simplicity of explanation, we set the number of points to be predicted for each sample to one in this context. The UES calculates the Mean Square Error (MSE) between the predicted probability distributions of M prediction heads and the reference distribution $\widetilde{p}_{i}$. The uncertainty $u^{S}_{i}$ for the $i$-th sample is as follow:
	
	\begin{equation}\label{eq:SampleUncertainty_Pose}
		u^{S}_{i}=\frac{1}{M}\sum_{m=1}^{M}\text{MSE} \big(p_{(i,m)}, \widetilde{p}_{i}\big),
	\end{equation}
	where $\text{MSE} \big( B, C \big)$ calculates the MSE between predicted probability distributions $B$ and $C$. 
	
	The $i$-th sample's weight $w^{S}_{i}$ can be specifically represented as:
	
	For a batch of unlabeled data $\mathcal{D}_{U}$, the sample weights are calculated from sample uncertainties $\mathcal{U}^{S}=\{u^{S}_{i}\}^{\mu N_B}_{i=1}$, as follows: 
	
	\begin{equation}\label{eq:SimpleWeight_pose}
		w^{S}_{i}=\frac{1}{\frac{u^{S}_{i}}{\text{MAX}(\mathcal{W}^{S})} + 1},
	\end{equation}
	where the function $\text{MAX}(A)$ returns the maximum value of set $A$.
	
	\begin{figure}[h!]
		\centering
		\includegraphics[width=3in]{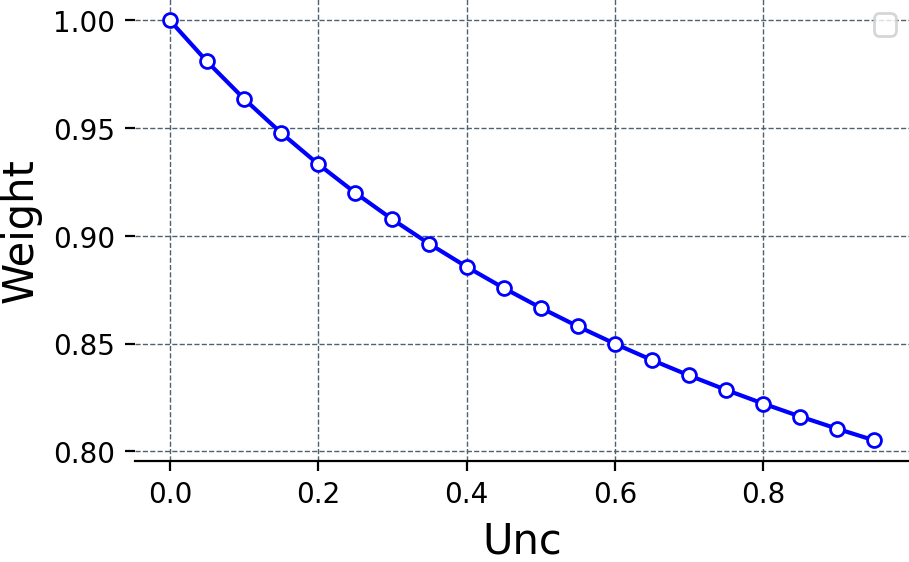}
		\caption{The long-tailed weights from Eq.~\eqref{eq:SimpleWeight_pose} give each sample a non-zero weight. \label{fig:long_tailed} }
	\end{figure}
	
	\textbf{Long-tailed weight.} As shown in Fig.~\ref{fig:long_tailed}, Eq.~\eqref{eq:SimpleWeight_pose} transforms uncertainty into long-tailed weights, ensuring that even samples with the highest uncertainty receive a weight greater than zero, thus guaranteeing the participation of all samples in the training process. The objective of this design strategy is to promote the involvement of a larger number of samples in model training. Specifically, during the initial stages of semi-supervised learning, incorporating more samples (including those predicted incorrectly) into the training process aids the model in swiftly capturing initial data features. As the training epochs increase, the model's overall predictive ability for all unlabeled samples improves, with accuracy gradually rising and the introduced noise progressively diminishing. This trend aligns with the need for more accurate supervisory information in the middle and late stages of training.
	
	\textbf{In classification} tasks, the UES quantifies the sample uncertainty of unlabeled data $x_i$ by calculating the average variance of each class across $M$ probability distributions predicted by $M$ prediction heads.
	
	For the sample $s_i$ of an unlabeled data $x_i$, the $M$ prediction heads predict $M$ probability distributions $\mathcal{P}_{i}=\{p_{(i,m)}\}^M_{m=1}$. For the $j$-th class of $\{c_{j}\}^{N_\mathcal{C}}_{j=1}$, where $N_C$ is the number of classes, the predictive probability of sample $s_i$ belonging to the $j$-th class by the $m$-th prediction head can be denoted as $\mathcal{C}_{(i,m,j)}$.
	
	The mean probability $\overline{\mathcal{C}}_{(i,j)}$ of $\mathcal{C}_{(i,m,j)}$ is given by:
	
	\begin{equation}\label{eq:SampleUncertainty_ClassMeanProbability}
		\overline{\mathcal{C}}_{(i,j)}=\frac{1}{M}\sum_{m=1}^{M}\mathcal{C}_{(i,m,j)}
	\end{equation}
	
	The variance $\text{var}_{(i,j)}$ of $\mathcal{C}_{(i,m,j)}$ is then calculated as:
	
	\begin{equation}\label{eq:SampleUncertainty_ClassVariance}
		\text{var}_{(i,j)}=\frac{1}{M}\sum_{m=1}^{M}\big(\overline{\mathcal{C}}_{(i,j)}-\mathcal{C}_{(i,m,j)}\big)^2
	\end{equation}
	
	To assess the uncertainty $u^{S}_i$ of the sample $s_i$, the average variance across all categories is calculated as:
	
	\begin{equation}\label{eq:SampleUncertainty_ClassFullVariance}
		u^{S}_{i}=\frac{1}{N_C}\sum_{j=1}^{N_C}\text{var}_{(i,j)}
	\end{equation}
	
	For a batch of unlabeled data $\mathcal{D}_{U}$, the sample uncertainties can be expressed as $\mathcal{U}^{S}=\{u^{S}_{i}\}^{\mu N_B}_{i=1}$. The sample weights $\mathcal{W}^{S}=\{w^{S}_{i}\}^{\mu N_B}_{i=1}$ are calculated by taking the reciprocal of the normalized sample uncertainties. This process is same as defined in  Eq.~\eqref{eq:SimpleWeight_pose}.
	
	\subsection{Prediction Head Uncertainty}
	\label{ssec:Prediction_Head_Uncertainty}
	
	In general, ensemble predictions exhibit a significant improvement in accuracy compared to individual base models. Motivated by this observation, for a batch of samples, we calculate the simple average probability distribution $\widetilde{\mathcal{P}}=\{\widetilde{p}_{i}\}^{\mu N_B}_{i=1}$ from multiple predictions generated by $M$ prediction heads, and use $\widetilde{\mathcal{P}}$ as a reference distribution for computing uncertainty. Here, $\widetilde{p}_{i}=\frac{1}{M}\sum_{m=1}^{M}p_{(i,m)}$, where $N_B$ denotes the batch size of the labeled data, and $\mu$ represents the ratio of the unlabeled data number to the labeled data number.
	
	\textbf{In regression} tasks, for simplicity of explanation, we set the number of points to be predicted for each sample to one in this context. The UES quantifies uncertainty by calculating the Mean Square Error (MSE) between the predicted probability distributions of each prediction head and the reference distribution $\widetilde{p}_{i}$. Specifically, the uncertainty $u^{H}_{m}$ of the $m$-th prediction head is given by:
	
	\begin{equation}\label{eq:HeadUncertainty_Pose}
		u^{H}_{m}=\frac{1}{\mu N_B}\sum_{i=1}^{\mu N_B}\text{MSE} \big(p_{(i,m)}, \widetilde{p}_{i}\big),
	\end{equation}
	where $\text{MSE} \big( B, C \big)$ calculates the MSE between the predicted probability distributions $B$ and $C$.
	
	Subsequently, to suppress erroneous predictions from individual prediction heads, we employ the softmax function to transform uncertainty $\mathcal{U}^{H}=\{u^{H}_{m}\}^{M}_{m=1}$ into sample weights $\mathcal{W}^{H}=\{w^{H}_{m}\}^{M}_{m=1}$. This process is as follows:
	
	\begin{equation}\label{eq:HeadNormalization_Pose}
		\mathcal{W}^{H}=\text{SOFTMAX}(-\mathcal{U}^{H})
	\end{equation}
	
	\textbf{In classification} tasks, the UES quantifies the uncertainty of the $m$-th prediction head by calculating the count of predictions whose predicted classes not align with the predicted class of the reference distribution $\widetilde{p}_{i}$. Here, $\widetilde{p}_{i}=\frac{1}{M}\sum_{m=1}^{M}p_{(i,m)}$. Specifically, the uncertainty $u^{H}_{m}$ of the $m$-th prediction head is given by:
	
	\begin{equation}\label{eq:HeadUncertainty_Classification}
		u^{H}_{m}=\frac{1}{\mu N_B}\sum_{i=1}^{\mu N_B}\sigma \big(\text{CLASS}(p_{(i,m)}) \neq \text{CLASS}(\widetilde{p}_{i})\big),
	\end{equation}
	where the function $\text{CLASS}(A )$ outputs the class index with the maximum probability of the distribution $A$. The function $\sigma(B)=1$ if the condition $B$ is true, and $\sigma(B)=0$ otherwise. $N_B$ and $\mu$ represent the batch size of the labeled data and the ratio of the unlabeled data number to the labeled data number, respectively.
	
	Subsequently, the prediction head weights $\mathcal{W}^{H}$ are normalized from $\mathcal{U}^{H}=\{u^{H}_{m}\}^{M}_{m=1}$. using the softmax function. This process is as follows:
	
	\begin{equation}\label{eq:HeadNormalization_Classification}
		\mathcal{W}^{H}=\text{SOFTMAX}(M-\mathcal{U}^{H}),
	\end{equation}
	where $M$ is the number of prediction heads.
	
	Given that predictions from prediction heads for unlabeled data can easily exhibit fluctuations, it is necessary to adopt the EMA method to smooth the weights of the prediction heads, in order to prevent such fluctuations from causing dramatic impacts on the calculated weights.
	
	\subsection{Application to SSL}
	\label{ssec:Application_to_SSL}
	
	For a batch of labeled data $\{(x_i,y_i)\}^{N_B}_{i=1}$, the supervised training is guided by the ground truth $y_i$ as follows:
	\begin{small}
		\begin{equation}\label{eq:SupevisedLoss}
			L_{l}=\frac{1}{N_B}\sum_{i=1}^{N_B}\frac{1}{M}\sum_{m=1}^{M}\text{CE}\big(p_{(i,m)}, y_{i}\big) ,
		\end{equation}
	\end{small}
	where $N_B$ is the batch size of the labeled data, and $\text{CE}(\cdot,\cdot)$ denotes the cross-entropy function. 
	
	For unlabeled data $\{x_i\}^{\mu N_B}_{i=1}$, the threshold $\tau$ is used to filter the unreliable predictions from the ensemble prediction $\overline{\mathcal{P}}_i$. The ensemble prediction $\overline{\mathcal{P}}_i$ is calculated as a weighted average:
	\begin{equation}\label{eq:PseudoLabel}
		\overline{\mathcal{P}}_i=\frac{1}{M}\sum_{m=1}^{M}\Gamma(\text{max}(w^{H}_{m} \cdot p_{(i, m)}) >\tau)\cdot w^{H}_{m} \cdot p_{(i, m)},
	\end{equation}
	where $w^{H}_{m}$ is the prediction head weight of the $m$-th prediction head, and $\Gamma(\cdot>\tau)$ is the indicator function for the confidence threshold $\tau$.
	
	$\overline{\mathcal{P}}_i$ in Eq.~\eqref{eq:PseudoLabel} is then used as a pseudo-label to supervise $\mathcal{P}_i$. Consequently, the ensemble supervised loss $L_{e}$ is as follows:
	\begin{equation}\label{eq:EnsembleLoss}
		L_{e}=\frac{1}{\mu N_B}\sum_{i=1}^{\mu N_B}\frac{1}{M}\sum_{m=1}^{M} w^{S}_{i} \text{CE}\big(p_{(i,m)},\overline{\mathcal{P}}_i\big),
	\end{equation}
	where $w^{S}_{i}$ is the sample weight of the $i$-th sample, and $\mu$ is the ratio of the number of unlabeled data number to that of labeled data number.
	
	\section{Experiments on Semi-supervised Pose Estimation}
	\label{sec:Experiments_on_Pose_Estimation}
	
	SSL 2D pose estimation is employed to validate the effectiveness of our method on the regression task.
	
	\subsection{Dataset and Evaluation Metrics}
	\label{ssec:Pose_Estimation_Dataset}
	
	\begin{figure}[h]
		\centering
		\subfloat{
			\includegraphics[width=1.1in]{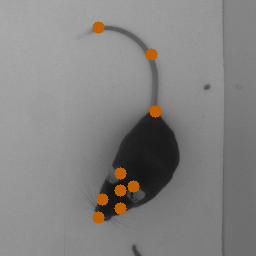}
			\includegraphics[width=1.1in]{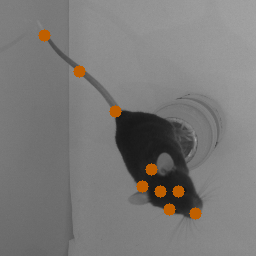}
			\includegraphics[width=1.1in]{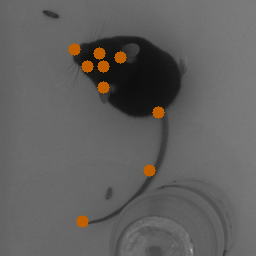}
		}
		\hfil
		\subfloat{
			\includegraphics[width=1.1in]{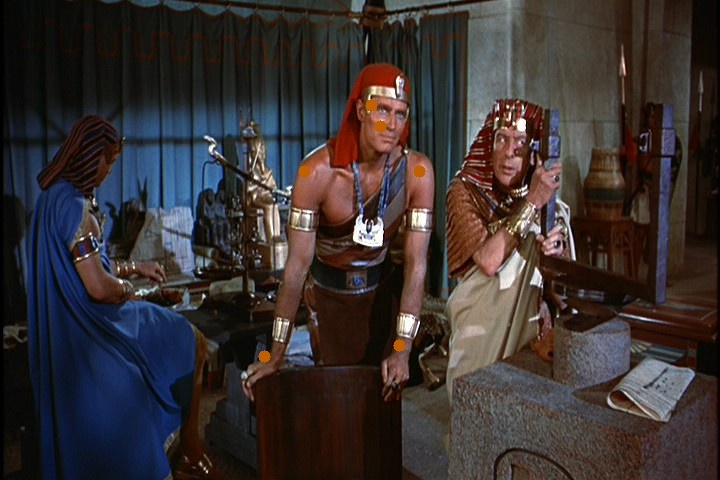}
			\includegraphics[width=1.1in]{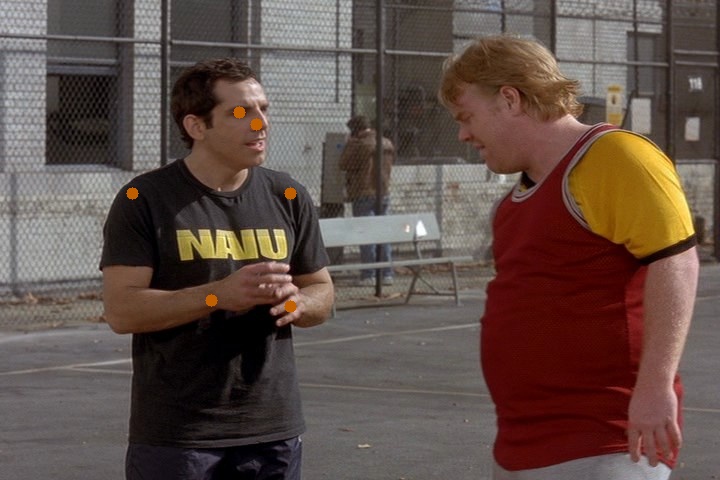}
			\includegraphics[width=1.1in]{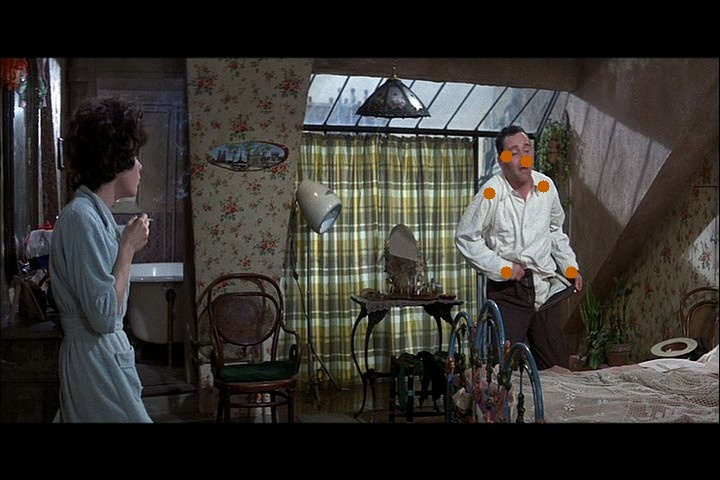}
		}
		\hfil
		\subfloat{
			\includegraphics[width=1.1in]{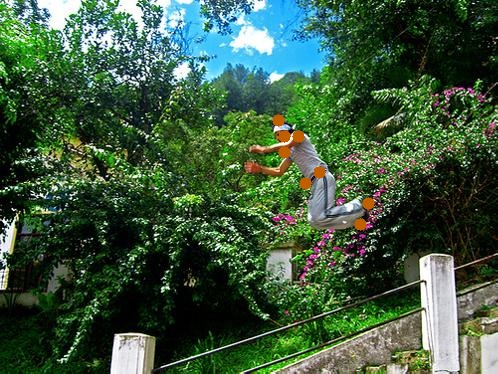}
			\includegraphics[width=1.1in]{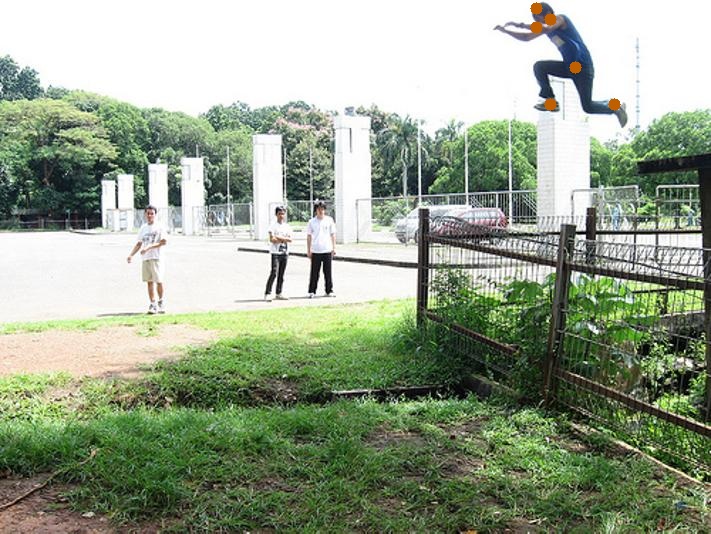}
			\includegraphics[width=1.1in]{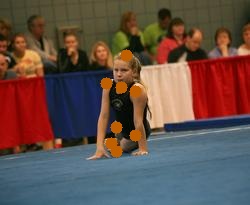}
		}
		\caption{Samples from the Sniffing (top row), FLIC (middle row), and LSP (bottom row) datasets are presented. \label{fig:D_Sample}}
	\end{figure}
	
	Our experiments are primarily conducted on three publicly available datasets: FLIC~\cite{sapp2013modec}, LSP~\cite{johnson2011learning}, and the Sniffing dataset\footnote{\url{https://github.com/Qi2019KB/Sniffing-Dataset}}. Fig.~\ref{fig:D_Sample} depicts the image data and their corresponding labels from the Sniffing dataset.
	
	To rigorously assess the performance of UES, we report the metrics of Mean Squared Error (MSE) and Percentage of Correct Keypoints (PCK). Specifically, the mean MSE is defined as follows:
	
	\begin{equation}\label{eq:MSE_formula}
		MSE =  \frac{1}{I} \frac{1}{K} \sum_{i=1}^{I} \sum_{k=1}^{K} \sqrt{|{pred}_{i,k} - gt_{i, k}|^{2}} ,
	\end{equation}
	where $I$, $K$ denote the number of images and keypoints, respectively. $i$ and $k$ are the indices of images and keypoints, respectively. ${pred}_{i,k}$ and $gt_{i,k}$ represent the predicted and ground-truth locations of the $k$-th keypoint in the $i$-th image.
	
	Mean Squared Error (MSE) serves as an intuitive metric for quantifying the Euclidean distance between predicted and ground-truth points. A lower MSE value indicates a smaller overall error. The mean of the Percentage of Correct Keypoints (PCK) is defined as follows:
	
	\begin{equation}\label{eq:PCK_formula}
		PCK = \frac{\sum_{i=1}^{I} \sum_{k=1}^{K} \sigma \big (\frac{d_{i,k}}{l} \leq T \big)}{\sum_{i=1}^{I} \sum_{k=1}^{K} 1} ,
	\end{equation}
	where $l$ is a normalized distance defined by the protocols of the different datasets. $T$ is a threshold, where $0\leq T \leq 1$, and the function $\sigma(A )$ is defined as $\sigma(A)=1$ if the condition $A$ is true, and $\sigma(A)=0$ otherwise.
	
	In contrast to MSE, PCK takes into account the variability in label errors across different datasets when evaluating the accuracy of pose estimation predictions. A higher PCK value indicates better accuracy of a method. PCK@$T$ denotes the use of a normalized distance $l$ to scale the MSE and the application of a threshold $T$ to determine prediction accuracy. Specifically, for the Sniffing dataset, $l$ represents the distance between the left and right eyes; for the FLIC dataset, $l$ is the distance between the left and right shoulders; and for the LSP dataset, $l$ is the distance between the left shoulder and left hip.
	
	We apply various PCK evaluation criteria across the three datasets, including PCK@0.1 ($T=0.1$), PCK@0.2 ($T=0.2$), PCK@0.3 ($T=0.3$), and PCK@0.5 ($T=0.5$), to assess the accuracy of keypoint predictions at different evaluation thresholds.
	
	\subsection{Configurations}
	\label{ssec:Pose_Estimation_Configurations}
	
	In all experiments, the pose model utilized was the Stacked Hourglass architecture with a stack number of 2, following the same configuration as described in~\cite{newell2016stacked}.
	
	All experiments were conducted using the Adam optimizer with a weight decay of 0.999. The learning rate was set to 0.00025, and the batch size was 4, with 2 labeled data points per batch. When calculating the prediction head weights, the decay value for the Exponential Moving Average (EMA)~\cite{tarvainen2017mean} was set to 0.7.
	
	SSL pose estimation experiments were conducted using DualPose as the SSL method. DualPose employed two data enhancement strategies to generate a set of sample pairs with varying levels of prediction difficulty. The strong data augmentation techniques included random rotation ($+/- 30$ degrees), random scaling ($ 0.75 - 1.25 $), and random horizontal flip. The weak data augmentation techniques included random rotation ($+/- 5$ degrees), random scaling ($ 0.95 - 1.05 $), and random horizontal flip.
	
	\subsection{Comparisons}
	\label{ssec:Pose_Estimation_Comparisons}
	
	\begin{table}[h]
		\caption{The MSE and PCK of SSL pose estimation with 200 training epochs are presented. ``Supervised'' denotes a pose model trained with only labeled data. ``X+CBE' and ``X+UES'' denote CBE and our UES network, respectively. ``X/Y'' denotes that X labeled data and Y unlabeled ones are used in SSL. \label{tab:Pose_Main_Results}}
		\centering
		\resizebox{\columnwidth}{!}{%
			\begin{tabular}{@{}lcccc@{}}
				\toprule 
				\multirow{3}{*}{Method} & \multicolumn{4}{c}{Sniffing} \\
				& \multicolumn{2}{c}{30/100} & \multicolumn{2}{c}{60/200} \\
				& MSE $\downarrow$ & PCK@0.2 (\%) $\uparrow$ & MSE $\downarrow$ & PCK@0.2 (\%)  $\uparrow$\\
				\midrule
				Supervised   & 4.72 $\pm$ 0.15              & 15.28 $\pm$ 2.52             & 4.60 $\pm$ 0.32              & 29.17 $\pm$ 1.73             \\
				\midrule
				\midrule
				DualPose~\cite{xie2021empirical} & 4.96 $\pm$ 0.09 & 16.67 $\pm$ 1.81     & 4.38 $\pm$ 0.07              & 30.56 $\pm$ 1.21             \\
				DualPose+CBE & \underline{3.59} $\pm$ 0.06 & \underline{67.36} $\pm$ 0.47 & \underline{4.15} $\pm$ 0.07  & \pmb{70.34} $\pm$ 0.35 \\
				DualPose+UES & \pmb{3.47} $\pm$ 0.11       & \pmb{70.83} $\pm$ 0.31       & \pmb{3.99} $\pm$ 0.09        & \underline{68.75} $\pm$ 0.25       \\
				\bottomrule
				\toprule 
				\multirow{3}{*}{Method} & \multicolumn{4}{c}{FLIC} \\
				& \multicolumn{2}{c}{50/100} & \multicolumn{2}{c}{100/200} \\
				& MSE $\downarrow$ & PCK@0.5 (\%) $\uparrow$ & MSE $\downarrow$ & PCK@0.5 (\%)  $\uparrow$\\
				\midrule
				Supervised   & 30.67 $\pm$ 2.07             & 54.17 $\pm$ 2.91             & 23.46 $\pm$ 1.86             & 65.62 $\pm$ 1.53             \\
				\midrule
				DualPose~\cite{xie2021empirical} & 30.99 $\pm$ 0.87 & 48.96 $\pm$ 0.84     & 21.05 $\pm$ 0.62             & 69.79 $\pm$ 0.32             \\
				DualPose+CBE & \underline{27.55} $\pm$ 0.42 & \underline{72.92} $\pm$ 0.93 & \underline{18.44} $\pm$ 0.56 & \underline{86.46} $\pm$ 0.51 \\
				DualPose+UES & \pmb{25.43} $\pm$ 0.27       & \pmb{80.21} $\pm$ 0.76       & \pmb{18.42} $\pm$ 0.43       & \pmb{87.50} $\pm$ 0.26       \\
				\bottomrule
				\toprule 
				\multirow{3}{*}{Method} & \multicolumn{4}{c}{LSP} \\
				& \multicolumn{2}{c}{100/200} & \multicolumn{2}{c}{200/300} \\
				& MSE $\downarrow$ & PCK@0.5 (\%) $\uparrow$ & MSE $\downarrow$ & PCK@0.5 (\%)  $\uparrow$\\
				\midrule
				Supervised   & 48.38 $\pm$ 1.57             & 16.41 $\pm$ 1.83             & 42.50 $\pm$ 1.06             & 32.81 $\pm$ 1.57             \\
				\midrule
				DualPose~\cite{xie2021empirical} & 44.43 $\pm$ 1.67 & 18.75 $\pm$ 2.31     & 37.64 $\pm$ 1.52             & 32.03 $\pm$ 1.33             \\
				DualPose+CBE & \underline{41.37} $\pm$ 1.08 & \underline{33.59} $\pm$ 1.54 & \pmb{34.95} $\pm$ 1.56 & \underline{56.25} $\pm$ 1.05 \\
				DualPose+UES & \pmb{37.72} $\pm$ 0.87       & \pmb{37.50} $\pm$ 1.21       & \underline{35.54} $\pm$ 1.12 & \pmb{58.59} $\pm$ 0.88       \\
				\bottomrule
			\end{tabular}%
		}
	\end{table}
	
	Table~\ref{tab:Pose_Main_Results} presents the performance of our method on three datasets: Sniffing, FLIC, and LSP. The results clearly demonstrate that our method outperforms the current state-of-the-art (SOTA) methods. Specifically, on the Sniffing dataset (containing 100 samples, with 30 labeled), DualPose+UES achieves a significant improvement of 3.47\% over DualPose+CBE on the rigorous PCK@0.2 metric. Similarly, on the FLIC dataset (containing 100 samples, with 50 labeled), DualPose+UES exhibits a remarkable performance boost of 7.29\% on the PCK@0.5 metric. Furthermore, on the LSP dataset (containing 200 samples, with 100 labeled), DualPose+UES also achieves a performance improvement of 3.91\% compared to DualPose+CBE. These substantial performance enhancements strongly evidence that the uncertainty-based evaluation mechanism employed by UES effectively guides the model to generate and select more accurate pseudo-labels, thereby significantly enhancing the performance of keypoint localization tasks.
	
	\begin{table}[h]
		\caption{Comparison of the performance of SSL methods for various PCK metrics. \label{tab:Pose_PCKs_Results}}
		\centering
		\resizebox{\columnwidth}{!}{%
			\begin{tabular}{@{}lccccc@{}}
				\toprule 
				\multirow{2}{*}{Method} & \multicolumn{5}{c}{Sniffing (30/100)} \\
				& PCK@0.5 (\%) $\uparrow$ & PCK@0.3 (\%) $\uparrow$ & PCK@0.2 (\%) $\uparrow$ & PCK@0.1 (\%) $\uparrow$ & MSE $\downarrow$ \\
				\midrule 
				DualPose~\cite{xie2021empirical} & 66.67 $\pm$ 1.25 & 46.53 $\pm$ 0.96     & 16.67 $\pm$ 1.81             & 0.00 $\pm$ 0.00              & 4.96 $\pm$ 0.09             \\
				DualPose+CBE & \underline{93.06} $\pm$ 0.41 & \underline{82.64} $\pm$ 0.45 & \underline{67.36} $\pm$ 0.47 & \underline{24.31} $\pm$ 1.03 & \underline{3.59} $\pm$ 0.06 \\
				DualPose+UES & \pmb{93.75} $\pm$ 0.24       & \pmb{86.11} $\pm$ 0.33       & \pmb{70.83} $\pm$ 0.31       & \pmb{27.08} $\pm$ 0.87       & \pmb{3.47} $\pm$ 0.11       \\
				\midrule
				\multirow{1}{*}{Method} & \multicolumn{5}{c}{FLIC (50/100)} \\
				\midrule
				DualPose~\cite{xie2021empirical} & 48.96 $\pm$ 0.84 & 36.46 $\pm$ 1.31     & 17.71 $\pm$ 1.65             & 0.00 $\pm$ 0.00              & 30.99 $\pm$ 0.87            \\
				DualPose+CBE & \underline{72.92} $\pm$ 0.93 & \underline{58.33} $\pm$ 0.82 & \underline{42.71} $\pm$ 1.05 & \underline{28.12} $\pm$ 1.23 & \underline{27.55} $\pm$ 0.42\\
				DualPose+UES & \pmb{80.21} $\pm$ 0.76       & \pmb{61.46} $\pm$ 0.78       & \pmb{51.04} $\pm$ 0.69       & \pmb{33.33} $\pm$ 0.81       & \pmb{25.43} $\pm$ 0.27      \\
				\midrule
				\multirow{1}{*}{Method} & \multicolumn{5}{c}{LSP (100/200)} \\
				\midrule
				DualPose~\cite{xie2021empirical} & 18.75 $\pm$ 2.31 & 10.16 $\pm$ 2.17     & 4.69 $\pm$ 1.85              & 0.00 $\pm$ 0.00              & 44.43 $\pm$ 1.67            \\
				DualPose+CBE & \underline{33.59} $\pm$ 1.54 & \underline{18.75} $\pm$ 1.51 & \underline{10.94} $\pm$ 0.85 & \underline{4.69} $\pm$ 1.34  & \underline{41.37} $\pm$ 1.08\\
				DualPose+UES & \pmb{37.50} $\pm$ 1.21       & \pmb{19.55} $\pm$ 1.21       & \pmb{11.77} $\pm$ 0.72       & \pmb{5.47} $\pm$ 1.03        & \pmb{37.72} $\pm$ 0.87      \\
				\bottomrule
			\end{tabular}
		}
	\end{table}
	
	Based on our empirical observations presented in Table \ref{tab:Pose_PCKs_Results}, the UES exhibits remarkable improvements in both the accuracy and stability of keypoint predictions. Specifically, on the Sniffing dataset (containing 100 samples, with 30 labeled), UES achieved enhancements of 0.69\%, 3.47\%, 3.47\%, and 2.77\% in PCK@0.5, PCK@0.3, PCK@0.2, and PCK@0.1, respectively, compared to the DualPose+CBE baseline. These improvements are consistently manifested across challenging datasets, such as FLIC and LSP. In particular, on the FLIC dataset (containing 100 samples, with 50 labeled), UES resulted in performance boosts of 7.29\%, 3.13\%, 8.33\%, and 5.21\% in PCK@0.5, PCK@0.3, PCK@0.2, and PCK@0.1, respectively, compared to DualPose+CBE. Similarly, on the LSP dataset (containing 200 samples, with 100 labeled), UES augmented DualPose by 3.91\%, 0.80\%, 0.83\%, and 0.78\% in PCK@0.5, PCK@0.3, PCK@0.2, and PCK@0.1, respectively.
	
	In summary, these substantial improvements demonstrate that our proposed method enhances the performance of existing SSL regression techniques by generating and selecting more accurate pseudo-labels, ultimately leading to superior keypoint prediction accuracy and stability.
	
	\subsection{Ablation}
	\label{ssec:Pose_Estimation_Ablation}
	
	\begin{table}[h]
		\caption{The results of the ablation experiment in pose estimation are presented, with 200 training epochs. Here, ``SW'' and ``PHW'' denote Sample Weights and Prediction Head Weights, respectively. \label{tab2_PoseAblationResults}}
		\centering
		\begin{tabular}{@{}lcccc@{}}
			\toprule 
			\multirow{2}{*}{Method} & \multirow{2}{*}{SW} & \multirow{2}{*}{PHW} & \multicolumn{2}{c}{FLIC (50/100)} \\
			& &          & MSE $\downarrow$        & PCK@0.5 (\%) $\uparrow$   \\
			\midrule
			Supervised   &            &            & 30.67       & 54.17       \\
			DualPose     &            &            & 30.99       & 48.96       \\
			DualPose+CBE &            &            & 27.55       & 72.92       \\
			\midrule
			DualPose+UES & \checkmark &            & 26.93       & 75.31       \\
			DualPose+UES & \checkmark & \checkmark & \pmb{25.43} & \pmb{80.21} \\
			\bottomrule
			\toprule 
			\multirow{2}{*}{Method} & \multirow{2}{*}{SW} & \multirow{2}{*}{PHW} & \multicolumn{2}{c}{LSP (100/200)} \\
			& &          & MSE $\downarrow$        & PCK@0.5 (\%) $\uparrow$   \\
			\midrule
			Supervised   &            &            & 48.38       & 16.41       \\
			DualPose     &            &            & 44.43       & 18.75       \\
			DualPose+CBE &            &            & 41.37       & 33.59       \\
			\midrule
			DualPose+UES & \checkmark &            & 40.08       & 35.10       \\
			DualPose+UES & \checkmark & \checkmark & \pmb{37.72} & \pmb{37.50} \\
			\bottomrule
			\toprule 
		\end{tabular}
	\end{table} 
	
	The ablation study results of our method on SSL pose estimation tasks are presented in Table~\ref{tab2_PoseAblationResults}.
	
	When using only SW, DualPose+UES achieves performance improvements of 2.39\% and 1.51\% on the FLIC dataset (containing 100 samples, with 50 labeled) and the LSP dataset (containing 200 samples, with 100 labeled), respectively, compared to DualPose+CBE. This result indicates that Sample Weights (SW) effectively filter out more accurate pseudo-labels.
	
	When both SW and Prediction Head Weights (PHW) are used simultaneously, DualPose+UES further achieves performance enhancements of 4.9\% and 2.4\% on these two datasets, respectively, compared to using only SW. This finding demonstrates that the weighted ensemble prediction guided by PHW can obtain more precise keypoint localization, thereby improving the quality of pseudo-labels and ultimately enhancing the training performance of the model.
	
	\section{Experiments on Semi-supervised Classification}
	\label{sec:Experiments_on_Classification}
	
	To demonstrate the enhancement achieved by combining the classical SSL method with our proposed UES, we evaluate the efficacy of our approach on the CIFAR-10/100 datasets. Specifically, FixMatch, a prominent example of classic weak-strong methods, represents the Pseudo-Labeling (PL) techniques in SSL technology. It is noteworthy that integrating the SSL method with our approach necessitates only two modifications: (1) incorporating our provided uncertainty calculation class library to transform the original ensemble prediction into a weighted average; and (2) utilizing sample weights in the computation of the loss function.
	
	\subsection{Configurations}
	\label{ssec:Classification_Configurations}
	
	In all experiments, the classification model employed was Wide ResNet, with a widen-factor of 2 for the CIFAR-10 dataset and 6 for the CIFAR-100 dataset, respectively.
	
	To ensure a fair comparison, the same hyperparameters from FixMatch were utilized across all experiments. Specifically, each experiment was conducted using the standard Stochastic Gradient Descent (SGD) optimizer with a momentum of 0.9, as suggested by~\cite{sutskever2013importance} and~\cite{polyak1964some}, and Nesterov momentum enabled, as introduced by~\cite{dozat2016incorporating}. The learning rate was set to 0.03. The batch size for labeled data was 32, while the batch size for unlabeled data was 224. Data augmentation for labeled data included random horizontal flipping and random cropping. For unlabeled data, weak data augmentation followed a similar process, but strong data augmentation incorporated RandAugment, as described by~\cite{cubuk2020randaugment}.
	
	To maximize the efficacy of long-tail weights, we adopted the following configuration strategies: In the initial five iteration rounds, we disabled the confidence threshold screening mechanism and relied solely on long-tail weights for sample selection and optimization. For the confidence thresholds applied in subsequent iterations, we set them lower than the values used in the baseline model to more fully integrate the influence of long-tail weights.
	
	Additionally, the number of multi-heads in both CBE and our proposed UES was set to 5.
	
	\subsection{Comparisons}
	\label{ssec:Classification_Comparisons}
	
	\begin{table}[h]
		\caption{The error rates (\%) of the SSL classification on CIFAR-10/100 datasets with 200 training epochs, where ``X+CBE'' and ``X+UES'' denote the CBE and our proposed UES network, respectively. \label{tab:Comparisons}}
		\centering
		\begin{tabular}{@{}lccc@{}}
			\toprule
			\multirow{2}{*}{Method} & \multicolumn{3}{c}{CIFAR-10} \\
			& 40 $\downarrow$ & 250 $\downarrow$ & 4000 $\downarrow$ \\
			\midrule
			
			FixMatch+CBE & 8.67 $\pm$ 0.53       & 6.55 $\pm$ 0.13       & 5.64 $\pm$ 0.15 \\
			
			FixMatch+UES & \pmb{8.47} $\pm$ 0.45 & \pmb{6.22} $\pm$ 0.11 & \pmb{5.54} $\pm$ 0.16 \\
			\bottomrule
			\multirow{2}{*}{Method} & \multicolumn{3}{c}{CIFAR-100} \\
			& 400 $\downarrow$ & 2500 $\downarrow$ & 10000 $\downarrow$ \\
			\midrule
			
			FixMatch+CBE & 52.47 $\pm$ 0.65 & 32.25 $\pm$ 0.43       & 23.86 $\pm$ 0.38 \\
			
			FixMatch+UES & \pmb{52.21} $\pm$ 0.53       & \pmb{31.92} $\pm$ 0.35 & \pmb{23.62} $\pm$ 0.31 \\
			\bottomrule
			\toprule
		\end{tabular}
	\end{table}
	
	As shown in Table~\ref{tab:Comparisons}, the UES demonstrates superior performance compared to current SOTA methods. Specifically, on the CIFAR-10 dataset, when using 40, 250, and 4000 labels, respectively, FixMatch+UES achieves performance improvements of 0.2\%, 0.33\%, and 0.1\% compared to FixMatch+CBE. Similarly, on the CIFAR-100 dataset, FixMatch+UES achieves performance enhancements of 0.26\%, 0.33\%, and 0.24\% when using 40, 250, and 4000 labels, respectively.
	
	\begin{figure}[h!]
		\centering
		\includegraphics[width=3in]{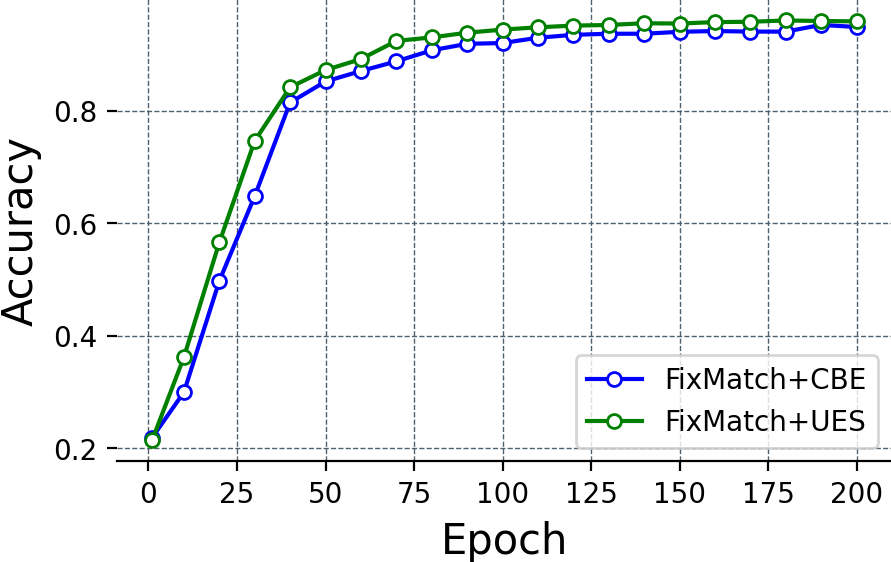}
		\caption{Comparison of the performance between CBE and UES, where UES employs long-tailed weights. \label{fig:long_tailed_accuracy} }
	\end{figure}
	
	The advantages of UES are primarily manifested in two aspects. Firstly, by calculating the uncertainty of samples, UES is able to select more reliable pseudo-labels. Secondly, by calculating the uncertainty of prediction heads and adopting a weighted ensemble approach, UES further enhances the accuracy of pseudo-labels, ultimately improving the model performance.
	
	Fig.~\ref{fig:long_tailed_accuracy} displays the performance curve of the FixMatch+UES model. Observations indicate that the performance of FixMatch+UES is consistently better than FixMatch+CBE throughout the entire training cycle. This advantage is primarily attributed to the long-tail weighting mechanism employed by UES, which models the utility of all unlabeled samples. Specifically, it assigns smaller weights to those pseudo-labels with lower reliability (i.e., higher uncertainty), allowing all pseudo-labels to participate in model training. This approach provides richer supervisory information for the model, thereby enhancing its robustness.

	\subsection{Ablation}
	\label{ssec:Classification_Ablation}
	
	\begin{table}[h]
		\caption{Results of ablation experiments on CIFAR-10 with 200 training epochs and 40 labels, where ``SW'' and ``PHW'' denote Sample Weights and Prediction Head Weights, respectively. \label{tab:Ablation}}
		\centering
		\begin{tabular}{@{}lccc@{}}
			\toprule 
			Method       & SW         & PHW         & Error rate(\%) $\downarrow$  \\
			\midrule
			FixMatch+CBE &            &            & 8.67      \\
			\midrule
			FixMatch+UES & \checkmark &            & 8.61      \\
			FixMatch+UES & \checkmark & \checkmark &\pmb{8.47} \\
			\bottomrule
			\toprule
		\end{tabular}
	\end{table}
	
	The ablation study results of our method on SSL classification tasks are presented in Table~\ref{tab:Ablation}.
	
	When using only SW, FixMatch+UES achieves a performance improvement of 0.06\% on the CIFAR-10 dataset (with 40 labels), compared to FixMatch+CBE. This result indicates that Sample Weights effectively filter out more accurate pseudo-labels, enhancing the training performance.
	
	When both SW and PHW are used simultaneously, FixMatch+UES further achieves performance enhancements of 0.14\% on the CIFAR-10 dataset (with 40 labels), compared to using only SW. This finding demonstrates that the weighted ensemble prediction guided by PHW can generate more accurate pseudo-labels.
	
	\section{Discussion}
	\label{sec:Analysis}
	
	\subsection{The Quality of Uncertainty in UES}
	\label{ssec:Decorrelation_Effect_Adapters}
	
	\begin{figure}[h]
		\centering
		\subfloat{\includegraphics[width=1.5in]{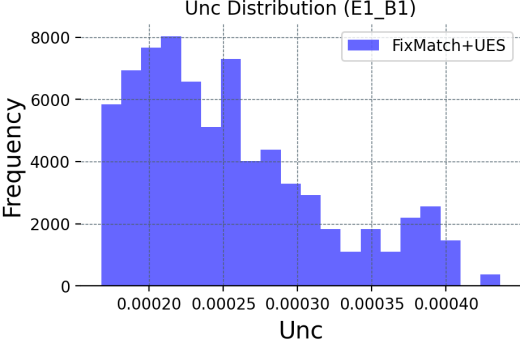}}
		\subfloat{\includegraphics[width=1.5in]{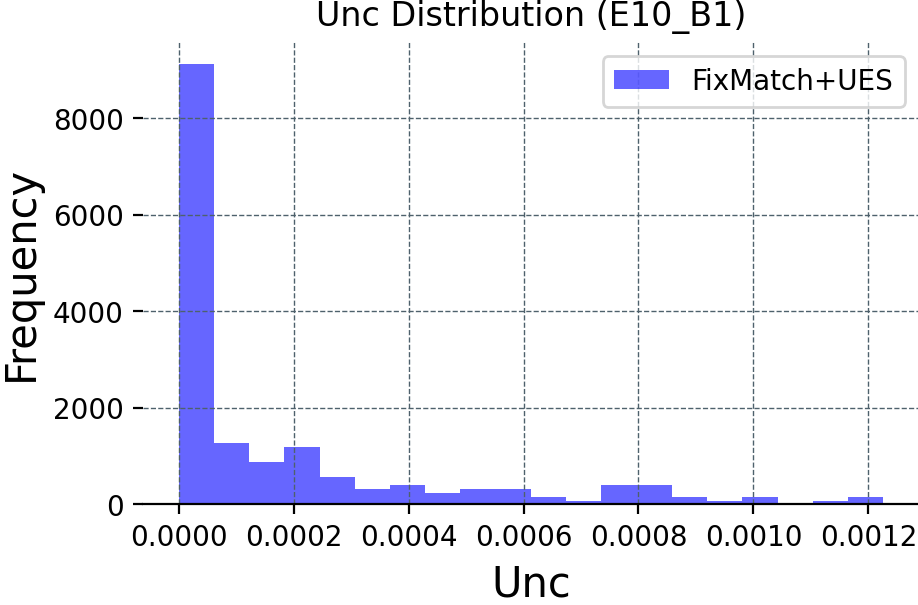}} \\
		\subfloat{\includegraphics[width=1.5in]{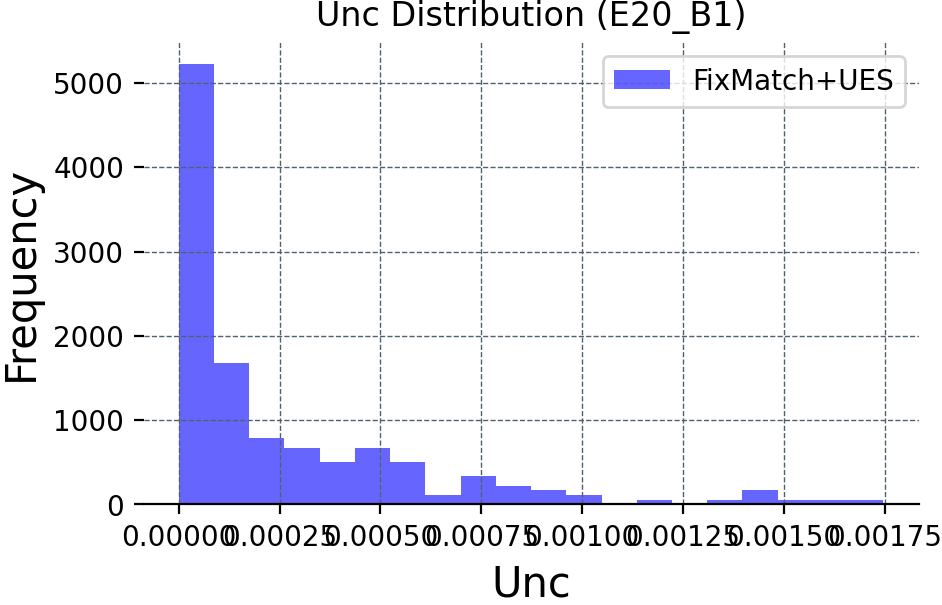}}
		\subfloat{\includegraphics[width=1.5in]{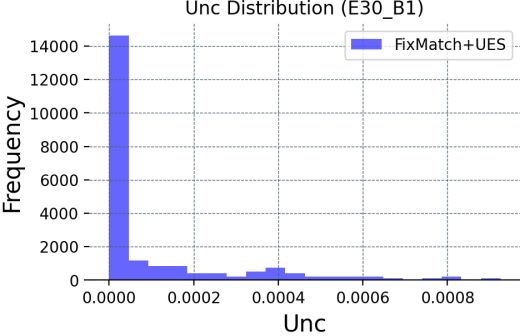}}
		\caption{The distribution of uncertainty for a batch of samples processed by the UES quantization method on the CIFAR-10 dataset, in the early stages of training. \label{fig:Unc_distribution}}
	\end{figure}
	
	\begin{figure}[h]
		\centering
		\includegraphics[width=2.5in]{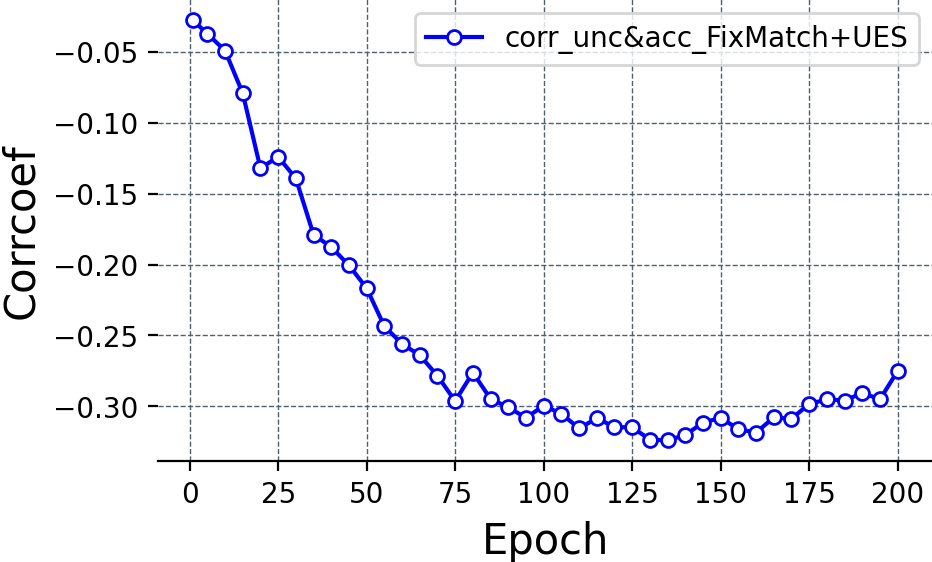}
		\caption{The correlation coefficient between the sample uncertainty quantized by UES and the accuracy of sample predictions. \label{fig:Unc_corrcoef}}
	\end{figure}
	
	The distribution of uncertainty quantified by UES for a batch of samples is shown in Fig.~\ref{fig:Unc_distribution}. As the training epochs increase, the quantified uncertainty gradually clusters near 0. However, an observation of Fig.~\ref{fig:Unc_corrcoef} reveals that the correlation between uncertainty and prediction accuracy is not high in the early stages of training, indicating that the quantification of uncertainty may not be accurate. Subsequently, as the training epochs increase, the accuracy of uncertainty gradually improves. This is because, in semi-supervised learning, due to the lack of label information, model predictions on unlabeled samples often face issues of accuracy and stability in the early stages of training, leading to potential risks of excessive prediction discrepancies among base models and overfitting.
	
	\section{Conclusions and Future Work}
	\label{sec:Conclusions_and_Future_Work}
	
	This paper proposes UES, a lightweight and architecture-agnostic ensemble structure. The UES quantifies the uncertainty of ensemble predictions for samples by assessing the consistency between ensemble predictions and the individual predictions of each prediction head within the ensemble structure. Additionally, UES calculates the prediction uncertainty of each prediction head and uses it as a weight when computing the ensemble prediction, optimizing the ensemble prediction results through this strategy. Experimental results demonstrate that our method achieves superior performance on classification datasets such as CIFAR-10/100, as well as on pose estimation datasets including Sniffing, FLIC, and LSP.
	
	In fact, different stages of SSL exhibit distinct characteristics, and the mechanism in Eq.~\ref{eq:SimpleWeight_pose} lacks adaptability. In future work, we will focus on dynamic weights mechanisms suitable for SSL. 
	
	\bibliography{reference}
	\bibliographystyle{IEEEtran}

	\begin{IEEEbiography}[{\includegraphics[width=25mm,height=32mm,clip,keepaspectratio]{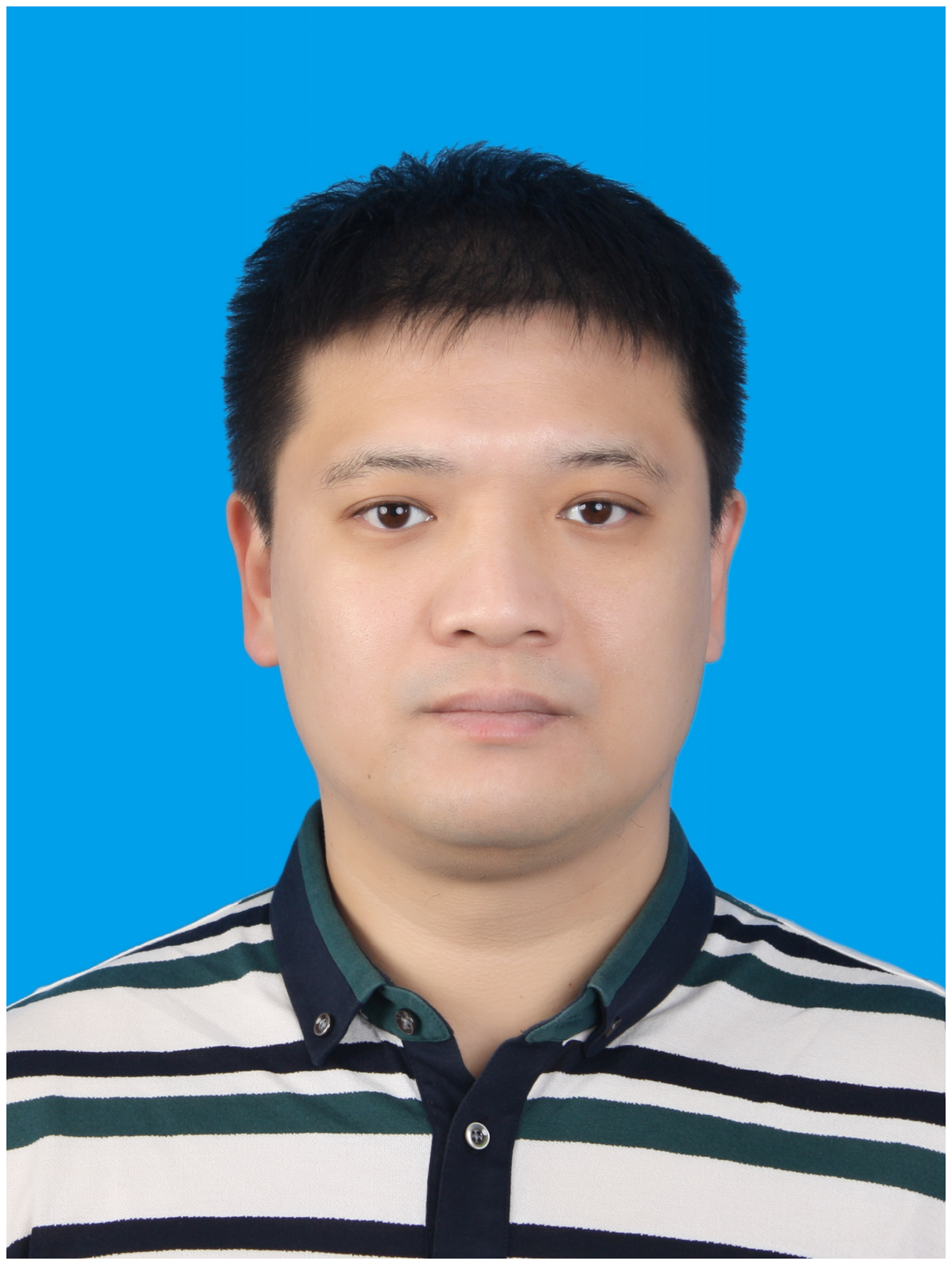}}]{Jiaqi Wu}
		He received his B.S. degree in Computer Science and Technology from Beijing University of Technology and his M.S. degree from Beihang University. Currently, he is pursuing a Ph.D. in Computer Science and Technology at Beijing University of Technology, with his primary research interests spanning computer vision, semi-supervised learning, and point detection.
	\end{IEEEbiography}

	\begin{IEEEbiography}[{\includegraphics[width=25mm,height=32mm,clip,keepaspectratio]{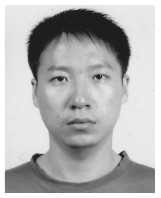}}]{Junbiao Pang}
		He received his B.S. degree and M.S. degree in computational fluid dynamics and computer science from Harbin Institute of Technology, Harbin, China, in 2002 and 2004, respectively, and his Ph.D. from the Institute of Computing Technology, Chinese Academy of Sciences, Beijing, China, in 2011.
		
		He is currently an Associate Professor with the Faculty of Information Technology, Beijing University of Technology (BJUT), Beijing, China. He has authored or coauthored approximately 20 academic papers in publications such as IEEE Transactions on Image Processing, ECCV, ICCV, and ACM Multimedia. His research interests include multimedia and machine learning for transportation applications.
	\end{IEEEbiography}
	
	\begin{IEEEbiography}[{\includegraphics[width=25mm,height=32mm,clip,keepaspectratio]{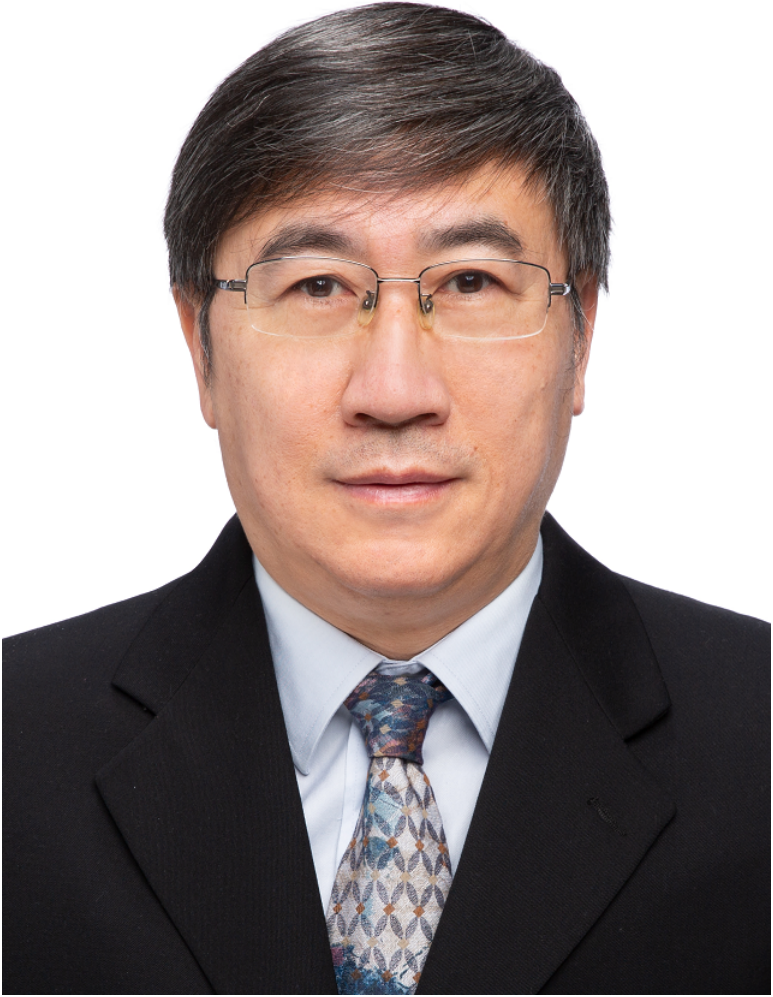}}]{Qingming Huang}
		He is a professor in the University of Chinese Academy of Sciences and an adjunct research professor in the Institute of Computing Technology, Chinese Academy of Sciences. His research areas include multimedia computing, image processing, computer vision and pattern recognition. He has authored or coauthored more than 400 academic papers in prestigious international journals and top-level international conferences. According to Google Scholar, these papers have been cited 13000+ times, with H-index 55. He was supported by the National Science Fund for Distinguished Young Scholars of China in 2010, deserved the Special Government Allowance of the State Council in 2011 and the National Hundreds and Thousands of Talents Project of China in 2014. He is the Fellow of IEEE and CCF. He is the associate editor of IEEE Trans. on CSVT and Acta Automatica Sinica, and the reviewer of various international journals including IEEE Trans. on PAMI, IEEE Trans. on Image Processing, IEEE Trans. on Multimedia, etc. He has served as general chair, program chair, track chair and TPC member for various conferences, including ACM Multimedia, CVPR, ICCV, ICME, ICMR, PCM, BigMM, PSIVT, etc.
	\end{IEEEbiography}

	\vfill
	
\end{document}